\pdfoutput=1

\documentclass[11pt]{article}

\usepackage[]{ACL2023}

\usepackage{times}
\usepackage{latexsym}
\usepackage{tikz}

\usepackage[T1]{fontenc}

\usepackage[utf8]{inputenc}

\usepackage{tcolorbox}
\usepackage{amsmath}
\usepackage{xcolor}

\usepackage{microtype}
 
\usepackage{inconsolata}
\usepackage{graphicx}
\usepackage{enumitem} 
\usepackage{amsmath} 
\usepackage{subcaption} 

\usepackage[table]{xcolor}  
\usepackage{colortbl}
\usepackage{array}
\usepackage{enumitem}
\usepackage{float}
\usepackage{wrapfig}

\definecolor{lightred}{RGB}{255,0,0}
\definecolor{lightgreen}{RGB}{0,255,0}
\definecolor{lightyellow}{RGB}{255,215,0}
\definecolor{lightorange_green}{RGB}{0,1,1}

\title{TOBUGraph: Knowledge Graph-Based Retrieval for Enhanced LLM Performance Beyond RAG}

\author{Savini Kashmira \\
  University of Michigan \\
  \texttt{savinik@umich.edu} \\\And
  Jayanaka L. Dantanarayana \\
  University of Michigan \\
  \texttt{jayanaka@umich.edu} \\ \And
  Joshua Brodsky \\
  University of Michigan \\
  \texttt{joshbrod@umich.edu} \\ \AND
 Ashish Mahendra \\
  Jaseci Labs \\
  \texttt{ashish.mahendra@jaseci.org} \\ \And
  Yiping Kang \\
  University of Michigan \\
  \texttt{ypkang@umich.edu} \\ \And
  Krisztián Flautner \\
  University of Michigan \\
  \texttt{manowar@umich.edu} \\ \AND
  Lingjia Tang \\
University of Michigan \\
  \texttt{lingjia@umich.edu} \\ \And
  Jason Mars \\
University of Michigan \\
  \texttt{profmars@umich.edu} \\
  }

\begin{document}
\maketitle

\begin{abstract}

Retrieval-Augmented Generation (RAG) is one of the leading and most widely used techniques for enhancing LLM retrieval capabilities, but it still faces significant limitations in commercial use cases. RAG primarily relies on the query-chunk text-to-text similarity in the embedding space for retrieval and can fail to capture deeper semantic relationships across chunks, is highly sensitive to chunking strategies, and is prone to hallucinations.
To address these challenges, we propose \textbf{TOBUGraph}, a graph-based retrieval framework that first constructs the knowledge graph from unstructured data dynamically and automatically. Using LLMs, TOBUGraph extracts structured knowledge and diverse relationships among data, going beyond RAG's text-to-text similarity. Retrieval is achieved through graph traversal, leveraging the extracted relationships and structures to enhance retrieval accuracy. This eliminates the need for chunking configurations while reducing hallucination. We demonstrate TOBUGraph’s effectiveness in \textbf{TOBU}, a real-world application in production for personal memory organization and retrieval. Our evaluation using real user data demonstrates that TOBUGraph outperforms multiple RAG implementations in both precision and recall, significantly enhancing user experience through improved retrieval accuracy.

\end{abstract}


\section{Introduction}


Integrating Large Language Models (LLMs) with external knowledge sources improves retrieval accuracy and enhances reliability  \cite{niu-etal-2024-ragtruth}. The state-of-the-art approach for such integration is Retrieval Augmented Generation (RAG) \cite{lewis2021retrievalaugmentedgenerationknowledgeintensivenlp, gao2024retrievalaugmentedgenerationlargelanguage}. 
Traditional RAG preprocesses documents by chunking text and storing the chunks in a vector database. During retrieval, it retrieves the top-ranked chunks based on vector similarity, and an LLM leverages those selected chunks to generate a response accordingly.

While traditional RAG-based approaches allow LLMs to incorporate external knowledge,  this methodology faces several key limitations:
\begin{itemize}[nosep]
\item RAG relies on query-chunk similarity in vector embeddings, comparing the query to each chunk individually without capturing broader contextual connections among text chunks.  However, in many domains, data can be interconnected. Failing to represent and leverage such relationships and structures beyond text-to-text similarity across multiple chunks often leads to low retrieval accuracy by RAG 
\cite{peng2024graphretrievalaugmentedgenerationsurvey}. 
\item Chunking and embedding strategies, such as chunk length and overlap size can significantly affect retrieval performance \cite{qu2024semanticchunkingworthcomputational}. 
\item When relevant chunks do not exist in the database for a given query, RAG may hallucinate \cite{10.1145/3703155}.
\end{itemize}

Indeed, our evaluation of RAG approaches using production data in a real-world application clearly highlights these limitations (Section~\ref{sec:eval}).


\begin{figure*}[!ht]
    \centering
    \begin{subfigure}[b]{0.495\textwidth}
        \includegraphics[width=\textwidth]{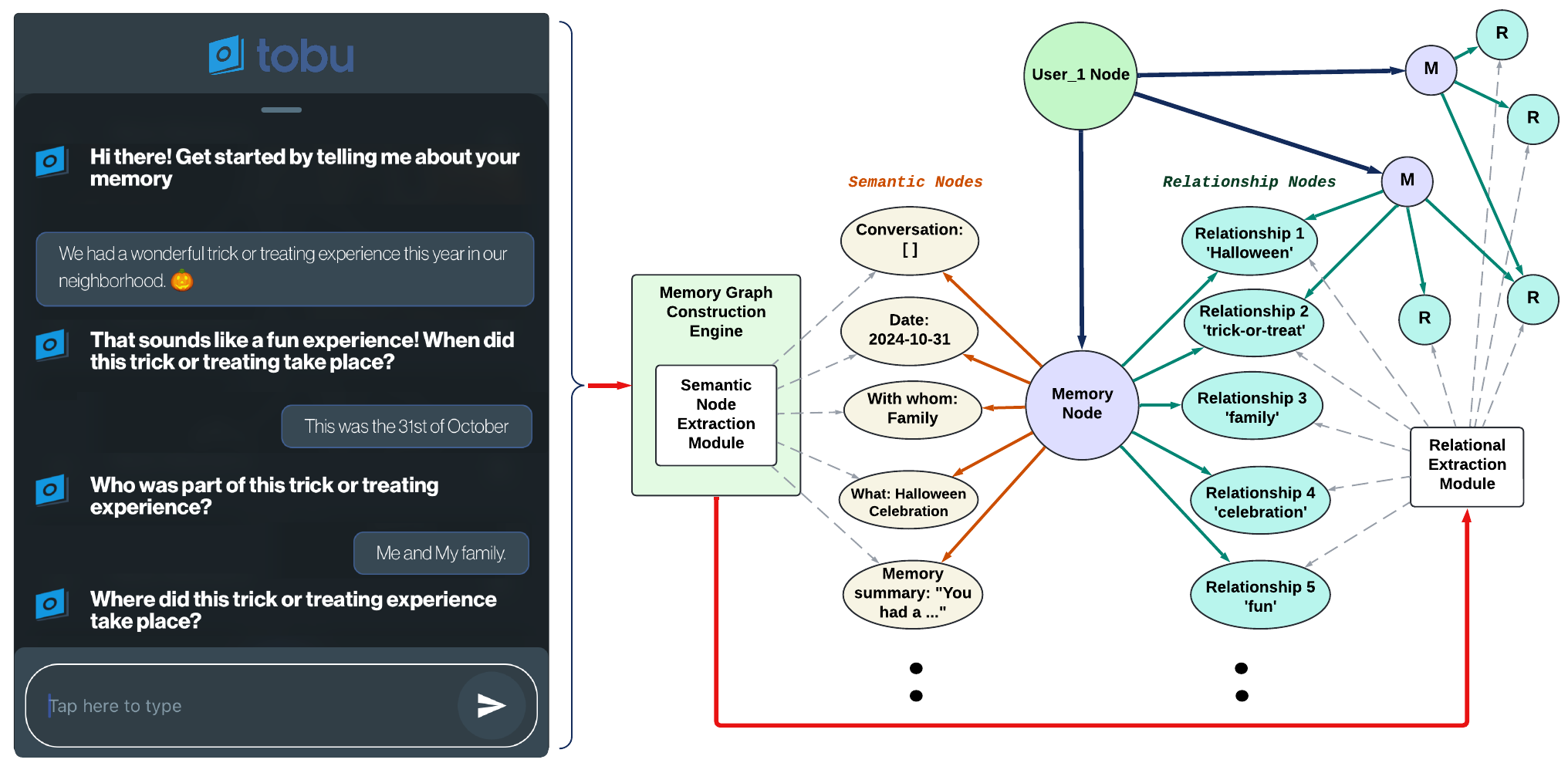}
        \caption{}
        \label{fig:TobuStructure}
    \end{subfigure}
    \begin{subfigure}[b]{0.495\textwidth}
        \includegraphics[width=\textwidth]{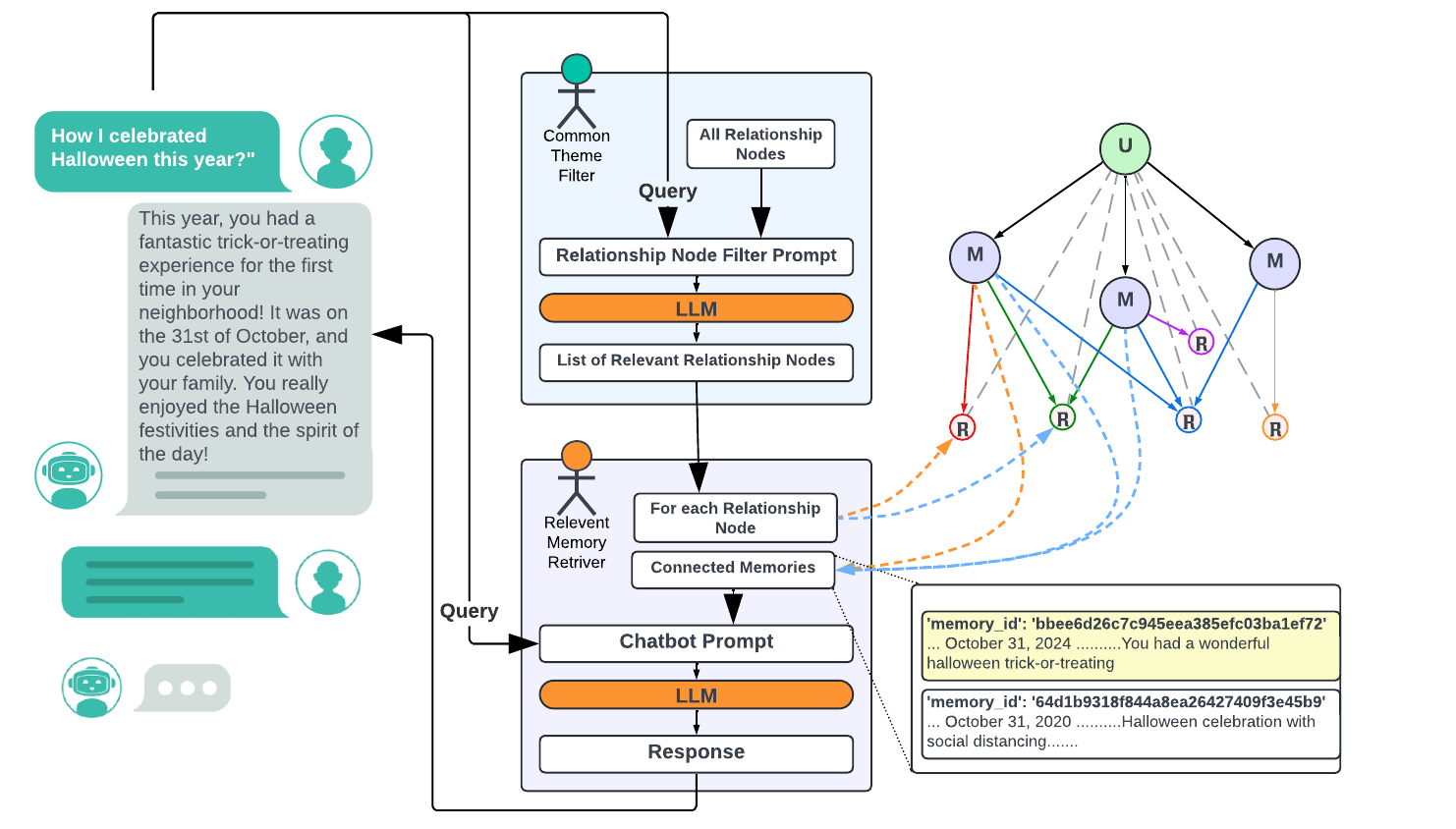}
        \caption{}
        \label{fig:TobuRetrieval}
    \end{subfigure}
    \caption{(a) Memory capturing workflow and (b) Memory retrieving workflow in TOBUGraph framework implemented in TOBU app.}
    \label{fig:tobu_system}
\end{figure*}

To address these limitations, it is important to uncover the relationships among unstructured data and leverage such relationships to improve retrieval performance. A promising approach is to structure data as knowledge graphs \cite{su2024knowledgegraphbasedagent,Hogan_2021}.
Prior work \cite{jin2024graphchainofthoughtaugmentinglarge, Wu2024CoTKRCE} introduces a graph-augmented retrieval technique that uses LLM reasoning over the knowledge graph through a chain-of-thought process. While this approach enhances retrieval, it assumes the existence of a predefined knowledge graph and overlooks its construction, which remains labor-intensive and lacks adaptability to dynamic data \cite{Hofer_2024}.  Designing a holistic graph-based retrieval framework that enables automatic knowledge graph construction and graph-based retrieval that captures deeper semantic relationships remains an open challenge.

In this work, we propose TOBUGraph, a novel graph-based retrieval augmentation framework. TOBUGraph leverages LLMs to automatically construct a knowledge graph from unstructured data. Unlike traditional RAG that stores data chunks in a vector database and compares query-chunk's text similarly, TOBUGraph extracts structured knowledge and diverse relationships among data and represents the structures and connections of data in a graph. Our novel graph structure is composed of \textbf{semantic nodes}, representing the key semantic information of data chunks, and \textbf{relationship nodes}, to represent diverse semantic relationships between semantic nodes. During retrieval, TOBUGraph leverages relationship nodes to prune irrelevant data and prioritize the retrieval on highly relevant data, improving retrieval precision. By traversing the pruned graph of all relevant interconnected nodes, we mitigate the limitations of traditional chunking and ensure completeness and high recall for the retrieval.


We implemented TOBUGraph in a real-world application called TOBU, designed for storing and retrieving personal memories. We define "personal memory" as user-provided images and videos coupled with details, context and narratives around them. When users upload an image, TOBUGraph will first leverage a multimodal LLM to extract key details and generate a summary of the image. Users can provide more details and refinements through a conversational AI assistant. TOBUGraph constructs a knowledge graph of such memories and facilitates users to interact and query about them. 

Using real-world user data of the TOBU app, we evaluated TOBUGraph approach against multiple RAG baseline implementations. TOBUGraph consistently outperformed these baselines in retrieval accuracy, efficiency, and user experience, receiving higher preference ratings across diverse memory retrieval scenarios.

The main contributions of this paper are as follows.
\begin{enumerate} [nosep]
    \setlength{\itemsep}{0em}
    \item A novel approach to extracting structured knowledge and diverse relationships among unstructured data and representing the structures and connections of data in a graph. 
    \item A novel approach to leverage such a knowledge graph to enable a more effective and efficient retrieval mechanism.
    \item Applying TOBUGraph in a real-world application for personal memory organization and retrieval.
    \item A comprehensive evaluation against RAG systems using real-world user data. TOBUGraph achieves \textbf{93.74\%} \textbf{\textit{precision}} (vs. 89.23\% best baseline), \textbf{91.96\%} \textbf{\textit{recall}} (vs. 82.26\% best baseline), and \textbf{92.84\%} \textbf{\textit{F1-score}} (vs 85.56\% best baseline). Our user experience evaluation shows that whenever TOBUGraph appears as a response option, evaluators are 75\% likely to choose it over RAG baselines.
\end{enumerate}

We plan to open-source our dataset and experimentation for further study.

\section{TOBUGraph}

In this section, we introduce TOBUGraph, a novel graph-based approach for information capture and retrieval. TOBUGraph overcomes RAG limitations by structuring information in dynamic graph-based representations that effectively capture data relationships. We describe TOBUGraph's implementation in the TOBU app for personal memory capture and retrieval.

During capturing (Figure~\ref{fig:TobuStructure}), TOBUGraph uses an LLM to automatically extract semantics from user inputs, transforming them into context-rich memories. Our system establishes memory relationships, forming a structured and contextually relevant memory graph. During retrieval (Figure~\ref{fig:TobuRetrieval}), users interact with a conversational AI assistant to retrieve information about the memories. 


\subsection{Memory Input Data Collection} \label{subsec:memory_input}

Our system combines a multimodal LLM with a conversational AI assistant to help users effortlessly create memory entries (Figure \ref{fig:TobuStructure}). When users provide multimedia inputs, such as images or videos, the multimodal LLM applies object recognition, emotion detection, scene recognition, and geolocation estimation to extract contextual details including date, location, people, activities, and emotions. Based on these details, the LLM generates an initial summary. TOBU AI assistant then engages users in a conversation, gathering additional information and refining the extracted data as needed. The summary dynamically updates as users provide more input, reflecting the most accurate and enriched version of the memory.

\subsection{Memory Graph Construction}
The Memory Graph Construction Engine organizes extracted contextual details and generated summaries into a structured, graph-based representation (Figure \ref{fig:TobuStructure}). This process occurs in two stages: individual memory structuring and cross-memory relationship discovery, which identifies and connects related memories across the entire collection.

\paragraph{Individual Memory Structuring:}
Semantic Nodes Extraction Module processes extracted memory details such as date, location and summaries to construct a graph for each individual memory, where each semantic detail is stored in a dedicated \emph{semantic node}. These nodes link to a central \emph{memory node} representing the memory itself. The initial memory graph is represented as $G = (V, E)$ where $V = M \cup S$ contains memory nodes $M = \{m_1, m_2, ..., m_n\}$ and semantic nodes $S = \{s_1, s_2, ..., s_k\}$, with edge set $E \subseteq M \times S$ connecting memory nodes to their semantic nodes.

\paragraph{Cross-Memory Relationship Discovery:}

TOBUGraph connects related memories across the user’s entire collection forming a unified structure called the Relational Memory Graph (RMG) (Figure \ref{fig:TobuStructure}). Using LLMs, Relational Extraction Module analyzes each memory node with its connected semantic nodes to extract common themes such as hobbies, locations, activities, significant dates, or frequently mentioned people. For each identified theme, a unique \emph{relationship node} is created, connecting all relevant memory nodes that share this common theme. To ensure robustness, the module performs a normalization step where extracted relationship labels are compared against existing nodes, merging variations and correcting minor spelling inconsistencies to avoid creating duplicate relationship nodes for the same semantic concept. The resulting RMG therefore provides a consistent and unified representation where shared relationships serve as central access points enabling the system to access all memories linked to a specific concept or theme.

The RMG extends the individual memory graphs $G$ to $G' = (V', E')$ where $V' = M \cup S \cup R$ includes relationship nodes $R = \{r_1, r_2, ..., r_p\}$, and $E' \subseteq (M \times S) \cup (M \times R)$ connects memory and relationship nodes. For each memory node $m_i$, connected semantic nodes are $S_i = \{s_j | (m_i, s_j) \in E'\}$ and connected relationship nodes are $R_i = \{r_j | (m_i, r_j) \in E'\}$.



\subsection{Memory Retrieval}




TOBUGraph enables users to retrieve memories through an integrated conversational AI assistant that interacts with the user's RMG, as shown in Figure \ref{fig:TobuRetrieval}. When a user initiates a memory retrieval request, the system collects all relationship nodes in the RMG and uses an LLM to filter the most relevant ones according to the user’s request. The system then traverses the RMG to retrieve the memory nodes connected to the filtered relationship nodes, along with their semantic content, which are passed to the conversational AI assistant.

The conversational AI analyzes this retrieved content to generate targeted responses. If the user’s request provides sufficient detail without ambiguity for the LLM, the response is direct; otherwise, the AI requests clarification. As conversations progress, the LLM in the conversational AI continually filters out irrelevant memories, refining the retrieved content to provide more accurate and contextually relevant answers.

\textbf{Retrieval Process Formalization:} The process follows three steps: (1) \textit{Relationship relevance}: $f(q, R) \rightarrow R' \subseteq R$ identifies relevant relationship nodes $R'$ for query $q$. (2) \textit{Memory retrieval}: $g(R', G') \rightarrow M' \subseteq M$ retrieves memory nodes $M'$ connected to $R'$ in the RMG. (3) \textit{Response generation}: $h(q, M', S') \rightarrow r$ generates response $r$ where $S' = \cup S_i|m_i \in M'$.

\begin{table*}[htbp]
\centering
\caption{Comparison of Baseline RAG Implementation Variants.}
\label{tab:rag-comparison}
\resizebox{\linewidth}{!}{
\begin{tabular}{|p{2cm}|p{9cm}|p{8cm}|}
\hline
\multicolumn{3}{|l|}{\textbf{Notations:} $M$: set of memories, $m_i$: individual memory, $n$: total memories, $C$: set of chunks, $c_i$: individual chunk, $l$: fixed chunk length} \\
\hline
\multicolumn{1}{|c|}{\textbf{Baseline}} & \multicolumn{1}{c|}{\textbf{Input Source for RAG Database}} & \multicolumn{1}{c|}{\textbf{Chunking Strategy}}\\
\hline
RAGv1 & AI generated memory summaries discussed in Section \ref{subsec:memory_input}  & \textbf{Fixed-size chunks with specified overlap}\\
    &  & $c_i = \textit{split}(\textit{summary}(M))$, $|c_i| = l$ and $|C|>|M|$ \\
\hline
RAGv2 & Conversation between the user and AI assistant discussed & \textbf{One complete memory as a single chunk}\\
      & in section \ref{subsec:memory_input}.&    $c_i = \textit{conversation}(m_i),$ $|C| = |M| = n$ \\
\hline
RAGv3 & Memory summaries as in RAGv1. & \textbf{One complete memory as a single chunk}\\
      & & $c_i = \textit{summary}(m_i),$   $|C| = |M| = n$ \\
\hline
\end{tabular}
}
\end{table*}



\section{Evaluation}
\label{sec:eval}

\subsection{Baselines}
\label{subsec:baselines}


To evaluate our proposed TOBUGraph approach, we implement three versions of naive RAG systems using LangChain and ChromaDB as baseline approaches. 
Three implementations differ primarily in their chunking strategies and input data sources as represented in Table~\ref{tab:rag-comparison}. RAGv1 processes memory summaries by splitting them into fixed-size chunks with a defined overlap. RAGv2 takes a different approach by using user-assistant conversations instead of summaries, with each chunk containing one complete conversation. RAGv3 also operates on memory summaries, but each chunk corresponds to a single summary.
\vspace{-0.2 em}

\subsection{Dataset Construction}
\label{subsec:dataset}
Using real memory data from 20 highly active TOBU app users with extensive memory databases, we anonymized the data and created 80 unique memory retrieval requests. We then applied the TOBUGraph memory retrieval technique to process conversations and retrieve relevant memories for each request. For comparison, we used the same retrieval requests with baseline RAG approaches, employing their respective retrieval techniques.

\subsection{Quantitative Analysis}
\subsubsection{Retrieval Metrics Evaluation}

\definecolor{darkgreen}{rgb}{0.0, 0.4, 0.0}
\begin{table}[h!]
\caption{$Precision$, $Recall$ and $F1-Score$ with 95\% confidence intervals for the TOBUGraph approach and baseline methods.}
\label{tab:eval_metrices}
\resizebox{\columnwidth}{!}{
\begin{tabular}{|c|c|c|c|c|}
\hline
\textbf {} & \textbf{RAGv1} & \textbf{RAGv2} & \textbf{RAGv3} & \textbf{TOBU}\\ \hline
$Precision$ (\%) & 85.92  & 86.30  & 89.23 & \textcolor{darkgreen}{\textbf{93.75}}  \\ \hline
$Recall$ (\%) & 66.40 & 79.60 &  82.26 & \textcolor{darkgreen}{\textbf{91.96}}\\ \hline
$F1-Score$ (\%) & 74.53 & 82.88 & 85.56 & \textcolor{darkgreen}{\textbf{92.84}}\\ \hline
    \end{tabular}}
\end{table}


To evaluate TOBUGraph against the baseline approaches discussed in Section \ref{subsec:baselines}, we use standard information retrieval metrics: $Precision$, $Recall$, and $F1-score$ with 95\% confidence intervals calculated using the dataset described in Section \ref{subsec:dataset}. As shown in Table \ref{tab:eval_metrices}, TOBUGraph demonstrates significant performance improvement across all metrics, achieving the highest $Precision$, $Recall$, and $F1-score$. This indicates that TOBUGraph significantly outperforms RAGv1, RAGv2, and RAGv3 both in accurately retrieving relevant memories and avoiding irrelevant retrievals, with an average improvement of approximately 7\% in overall effectiveness in $F1-score$ compared to RAGv3, the next-best performing approach.

\renewcommand{\arraystretch}{1} 

\begin{table*}[h]
    {\fontsize{7}{9}\selectfont

    \caption{Four categorization levels of user requests based on memory retrieval complexity and nature.}
    \label{tab:retrivel_levels}
    \centering
    \begin{tabular}{| m{4cm} | m{6cm} | m{5cm} |}
        \hline
        \textbf{Category} & \textbf{Description} & \textbf{Example User Request} \\
        \hline
        \textbf{Level 1: Single memory retrieval} & Simple questions requiring direct retrieval of a single memory. No need for relationships between memories. & 
        \textit{"When did I have my first dinner with my boarding mates and how was that experience?"} \\
        \hline
        \textbf{Level 2: Linked memory retrieval} & Requires connecting two or three directly related memories to provide an answer. & 
        \textit{"What were the best parts of my hiking and rafting trips?"} \\
        \hline
        \textbf{Level 3: Multi-memory retrieval} & When many memories within the database are contextually relevant to formulating a response. & 
        \textit{"What are the activities we have done during the summer and fall?"} \\
        \hline
        \textbf{Level 4: Semantic or pattern-based memory retrieval} & Requires identifying patterns, trends, or deeper relationships involving different times, locations, or people from multiple memories. & 
        \textit{"Tell me about the memorable places I enjoyed with my friends, including James, over the past year."} \\
        \hline
    \end{tabular}}
\end{table*}
\vspace{-0.3em}
\subsubsection{User Experience Evaluation}



To quantitatively evaluate TOBUGraph against baseline methods, we conducted a human-based study using double-blind pairwise comparison via crowd-sourcing using the SLAM tool \cite{10590016}. In each evaluation, participants were presented with two responses for the same user request from two different approaches and asked to compare them side by side. A total of 480 evaluators each completed 10 comparisons, resulting in 4,800 pairwise evaluations. Responses from TOBUGraph and RAG baselines were each evaluated 1,200 times against all other approaches. 

\begin{figure}[ht]
    \centering
    \vspace{-0.3cm}
    \includegraphics[width=0.9\linewidth]{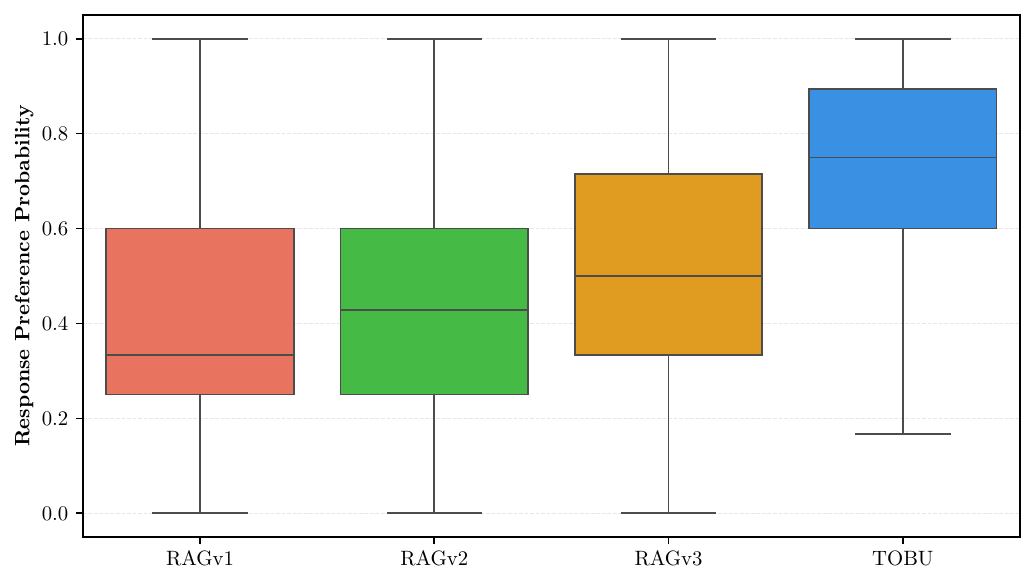}
        \caption{Distribution of evaluator preference of each approach, as probabilities. Among 480 human evaluators, TOBUGraph responses are preferred 75\% of the time on average, when present as a response option in a pairwise comparison. Furthermore, the preference distribution for TOBUGraph has lower variance, indicating more consistent performance compared to other approaches.}
    \label{fig:user_request_pref}
\end{figure}

We analyze evaluator preferences by measuring the probability of selecting each approach, as shown in Figure \ref{fig:user_request_pref}. Among the 480 evaluators, TOBUGraph responses are preferred 75\% of the time on average when presented as a response option in pairwise comparisons, significantly outperforming baseline methods. The distribution shows lower variance for TOBUGraph, indicating more consistent favorability. Among baselines, RAGv1 was least favored, while RAGv3 performed better than RAGv2 due to incorporating user-enriched memory summaries instead of conversations. These results highlight TOBUGraph’s effectiveness in delivering a more satisfying user experience.

\begin{figure}[ht]
    \centering
    \vspace{-0.3cm}
    \includegraphics[width=0.7\linewidth]{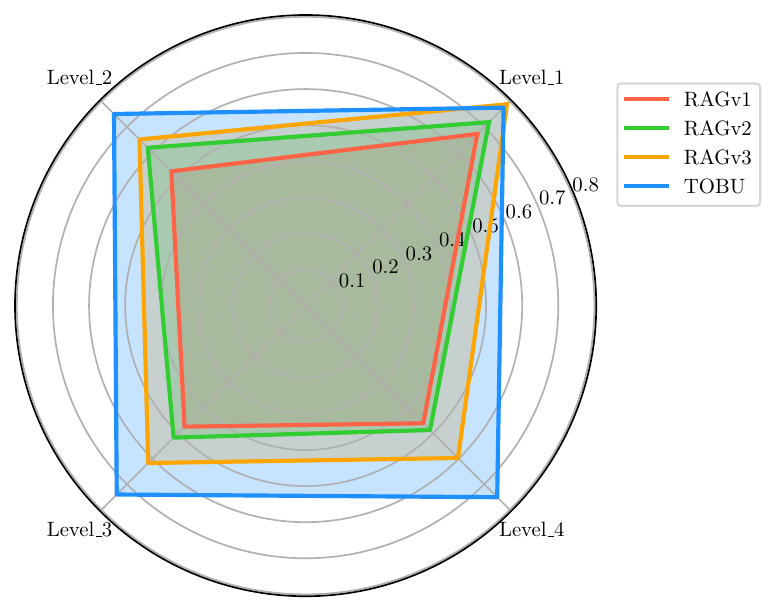}
        \caption{Evaluator preferences for each approach, measured as probabilities across four categorization levels based on memory retrieval complexity and nature. TOBUGraph consistently achieves the highest preference among evaluators across all levels, outperforming other approaches regardless of question complexity.}
    \label{fig:preference_distribution}
\end{figure}



To further analyze results, we categorized user requests of the dataset into four levels based on complexity and nature of the memory retrieval technique (Table \ref{tab:retrivel_levels}). Figure \ref{fig:preference_distribution} shows evaluator preferences across these levels.


For Level 1 user requests involving single memory retrieval, TOBUGraph and baseline RAG approaches perform similarly with nearly equal evaluator preference, since answering these questions does not require identifying relationships between multiple memories. As we progress to Levels 2, 3, and 4, the preference for RAG approaches declines due to the increasing complexity of memory retrieval (Figure \ref{fig:preference_distribution}). At Level 3, generating complete responses may require fetching a large proportion of the memory database, but RAG retrieves only top-$k$ relevant chunks, risking missing crucial context and leading to incomplete answers and a reduced user preference. Additionally, RAG embeddings prioritize text-to-text similarity, often failing to capture complex relationships between memories. This limitation affects memory retrieval at Level 4, decreasing user preference for RAG.

In contrast, TOBUGraph consistently maintains a higher preference across all levels, with an average selection rate of approximately 75\%. This strong performance is due to TOBUGraph’s ability to capture deeper semantic relationships through relationship nodes, enabling retrieval of highly relevant memories. These results indicate that, regardless of memory retrieval request complexity, TOBUGraph remains a highly effective solution, outperforming RAG-based methods.

\begingroup
\paragraph{\textbf{User Study Feedback}:} Users participated for evaluation further highlighted TOBUGraph's strengths, with comments such as \textit{"Response B (TOBU) has a smoother flow and includes five events, but Response A (RAGv2) only lists three events. Also, Response B describes each event more detail."}, \textit{"Response A (TOBU) is clearer and informative. It presents two Disney park visits and activities, But Response B (RAGv1) includes only one visit and incorrectly names the show they watched."} and \textit{"Response A (TOBU) provided a detailed narrative about two separate trips, while Response B (RAGv3) focused on a single trip but merged details from both."}
\endgroup

\subsection{Qualitative Analysis}






\begin{figure*}[!ht]
    \centering
    \begin{subfigure}[b]{0.66\textwidth}
        \includegraphics[width=\textwidth]{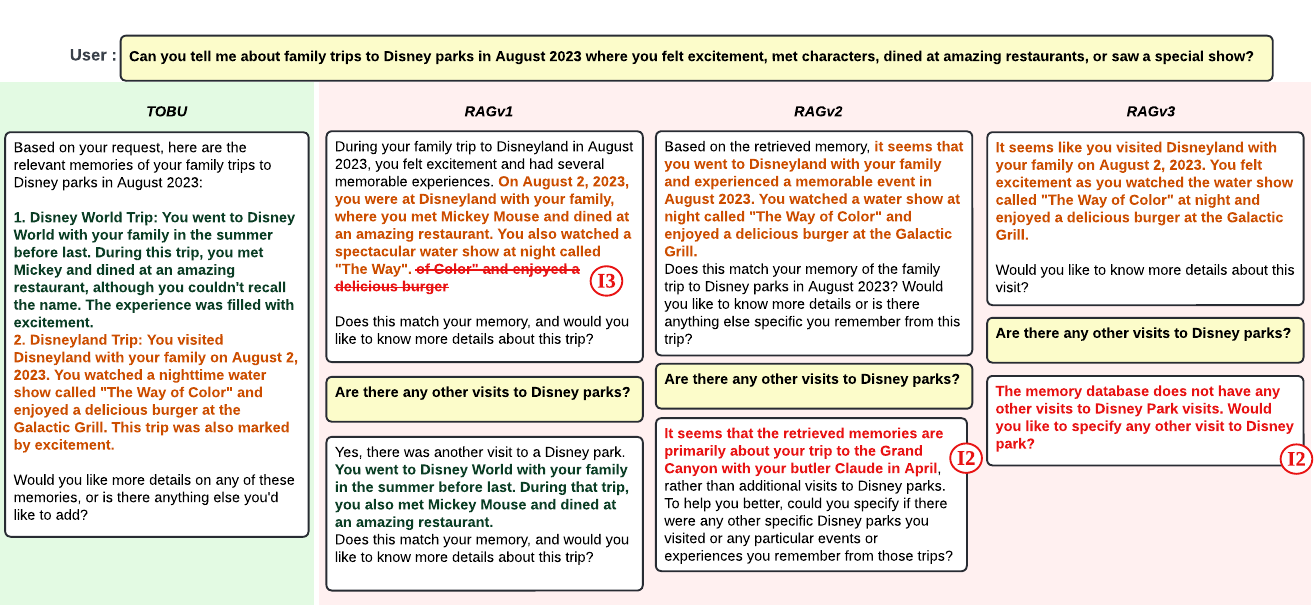}
        \caption{}
        \label{fig:Conv_1}
    \end{subfigure}
    \hfill  
    \begin{subfigure}[b]{0.238\textwidth}
        \includegraphics[width=\textwidth]{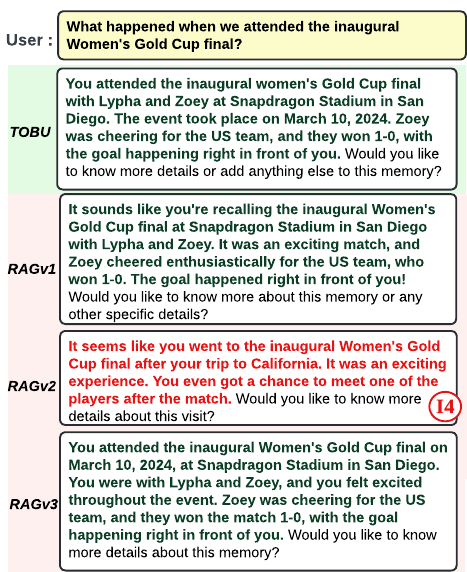}
        \caption{}
        \label{fig:Conv_3}
    \end{subfigure}
    \hfill  
    \begin{subfigure}[b]{0.66\textwidth}
        \includegraphics[width=\textwidth]{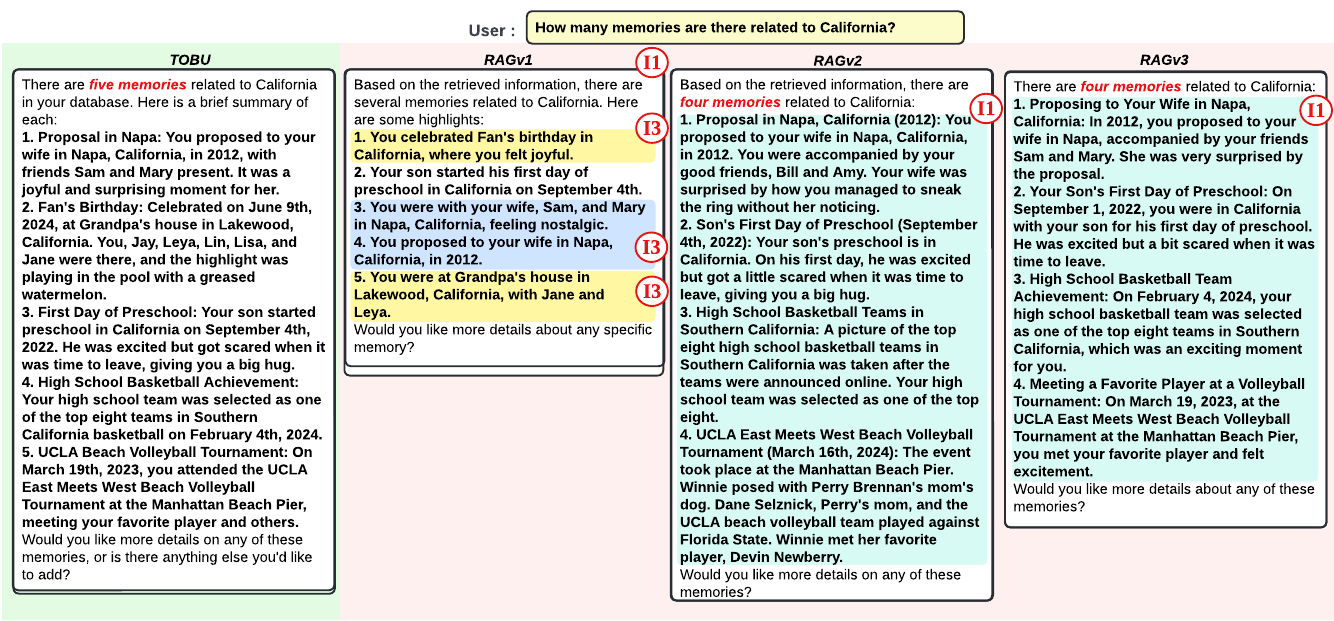}
        \caption{}
        \label{fig:Conv_2}
    \end{subfigure}
    \label{fig:conversations}
    \caption{Example conversations from the dataset discussed in section \ref{subsec:dataset} where (a) having issues \textbf{I1} and \textbf{I2}, (b) representing hallucination as in \textbf{I4} while (c) demonstrating issues \textbf{I2} and \textbf{I3} from Table \ref{tab:qualitative-comparison}.}
    \label{fig:conversations}
\end{figure*}

To evaluate TOBUGraph approach against baseline RAG models using the dataset created in Section \ref{subsec:dataset}, we also conducted a qualitative analysis. Key observations are summarized in Table \ref{tab:qualitative-comparison}, with detailed discussion below.

\begin{table}[h]
\caption{Qualitative comparison between memory retrieval techniques.}
\label{tab:qualitative-comparison}
\resizebox{\columnwidth}{!}{
\begin{tabular}{|p{9cm}|c|c|c|c|}
\hline
\textbf{Qualitative Observations}& \textbf{RAGv1} & \textbf{RAGv2} & \textbf{RAGv3} & \textbf{TOBU} \\
&  &  &  & \textbf{Graph} \\
\hline
(I1) Low recall due to top $k$ chunk limitation & \cellcolor{lightred} & \cellcolor{lightred} & \cellcolor{lightred} & \cellcolor{lightgreen} \\
\hline
(I2) Low recall due to embedding quality & \cellcolor{lightyellow} & \cellcolor{lightred} & \cellcolor{orange} & \cellcolor{lightgreen} \\
\hline
(I3) Splitting a memory into multiple chunks & \cellcolor{lightred} & \cellcolor{lightgreen} & \cellcolor{lightgreen} & \cellcolor{lightgreen} \\
\hline
(I4) Hallucinations during memory retrieval failures & \cellcolor{lightyellow} & \cellcolor{lightyellow} & \cellcolor{lightyellow} & \cellcolor{lightgreen} \\
\hline
\hline
\end{tabular}
}
\resizebox{\columnwidth}{!}{
\centering
\begin{tikzpicture}
    \shade[left color=red!100!white, 
           right color=green!100!white, 
           middle color=yellow!100!white] 
        (0,0) rectangle (10,0.2);

    \node[below] at (0,0) {Worst};
    \node[below] at (5,0) {Medium};
    \node[below] at (10,0) {Best};
\end{tikzpicture}
}
\end{table} 

\paragraph{(I1) Low recall due to top $k$ chunk limitation:}
Baseline RAG approaches retrieve only the top $k$ chunks, potentially missing relevant memories if their count exceeds $k$ (\textbf{I1} in Table~\ref{tab:qualitative-comparison}). As illustrated in Figure~\ref{fig:Conv_2}, TOBU retrieves all five relevant memories by leveraging graph traversal through the "California" relationship node, capturing all connected memories. In contrast, RAGv1, RAGv2, and RAGv3 fail to retrieve all the relevant memories. While RAGv1 appears as it retrieves five memories, two of them are redundant due to the splitting of a single memory, an issue further discussed in \textbf{I3}. Meanwhile, RAGv2 and RAGv3 retrieve only four relevant memories, omitting the memory labeled as `2' in TOBUGraph’s retrieval. 
\paragraph{(I2) Low recall due to embedding quality:}
RAG approaches rely on the quality of chunk embeddings for precise retrieval. However, in our use case, embedding quality declines as chunk length increases in the order of RAGv1, RAGv3 and RAGv2. This degradation affects retrieval performance, sometimes causing RAG methods to miss relevant memories (\textbf{I2} in Table \ref{tab:qualitative-comparison}). As illustrated in Figure \ref{fig:Conv_2}, TOBUGraph retrieves two related memories by traversing the graph via the "Disney" relationship node, without relying on any chunking or embedding strategies. In contrast, RAGv2 and RAGv3 retrieve only one memory, even after a follow-up query, omitting the "Disney World" memory. While RAGv1 retrieves both, it requires an additional follow-up question. 
\paragraph{(I3) Splitting a memory into multiple chunks:}
Unlike RAGv2 and RAGv3 that treat each memory as a single chunk, RAGv1’s chunking strategy unintentionally split memories (\textbf{I3} in Table \ref{tab:qualitative-comparison}). This can cause missing key details of a memory and misinterpreting a single memory as multiple distinct ones. In Figure \ref{fig:Conv_1} RAGv1, memories `3' and `4' originate from the same entry but are mistakenly treated as distinct, similar to `1' and `5'. Figure \ref{fig:Conv_2} further highlights this issue as the first memory retrieved by RAGv1 omits the correct name of the water show, as the strike-through content is absent in the response. In contrast, TOBUGraph avoids this issue entirely, as it employs the graph-based approach that preserves memory integrity without the need for chunking.


\paragraph{(I4) Hallucinations during memory retrieval failures:}
Baseline RAG models hallucinate when retrieval fails, fabricating information instead of returning valid entries (\textbf{I4}, Table~\ref{tab:qualitative-comparison}). Figure~\ref{fig:Conv_3} shows RAGv2 hallucinating because RAG relies on unstructured data, losing relationships between memories. In contrast, TOBUGraph structures memories as a graph, leveraging relationships for better retrieval. For example, when searching for the Women's Gold Cup final, TOBUGraph traverses through related relationship nodes, "inaugural", "Gold Cup", and "final" to retrieve relevant memories. This structured approach mitigates hallucinations by ensuring retrieval is based on existing relationships.


\vspace{-0.3 em}

\section{Related Works}
\vspace{-0.4 em}

Information retrieval with LLMs \cite{niu-etal-2024-ragtruth} often employs RAG, a state-of-the-art method~\cite{asai2023selfraglearningretrievegenerate, gao2024retrievalaugmentedgenerationlargelanguage, wu2024retrieval, 10.5555/3524938.3525306, karpukhin-etal-2020-dense}. However, RAG systems face several challenges: difficulty capturing deeper relationships between chunks beyond text-to-text similarities \cite{peng2024graphretrievalaugmentedgenerationsurvey}, sensitivity to chunking strategies \cite{qu2024semanticchunkingworthcomputational}, and hallucination risks \cite{sun2025redeepdetectinghallucinationretrievalaugmented, 10.1145/3703155}. 

Graph-based retrieval methods often address these issues~\cite{jin2024graphchainofthoughtaugmentinglarge, Wu2024CoTKRCE, hu2024graggraphretrievalaugmentedgeneration, su2024knowledgegraphbasedagent, chen2024unifiedtemporalknowledgegraph, zhang2022generative, peng2024graphretrievalaugmentedgenerationsurvey, 10504973, kim2024leveraging}. While knowledge graph construction is labor-intensive and struggles with dynamic data \cite{Hofer_2024}, \citeauthor{edge2025localglobalgraphrag} use LLMs to generate and update knowledge graphs primarily for creating summaries and RAG-based retrieval in GraphRAG, our approach retrieves information by traversing the graph.

\section{Conclusion}
\vspace{-0.3em}

In this paper, we introduce TOBUGraph, a novel framework that integrates LLM-powered knowledge graph construction with graph-based retrieval to enhance information retrieval while addressing RAG limitations. TOBUGraph improves retrieval accuracy by capturing deeper semantic relationships between entries. This approach is implemented in a mobile application called TOBU for memory retrieval. Our evaluation using real-world data from the TOBU database demonstrates that TOBUGraph consistently outperforms RAG baselines in precision, recall, and user preference ratings, highlighting its effectiveness in real-world scenarios.


\section*{Limitations}


While TOBUGraph demonstrates strong performance on the TOBU dataset, our evaluation is limited to users with relatively modest-sized memory collections. In real-world scenarios, users may accumulate thousands of memories, and scaling to such large collections presents new challenges. Specifically, graph construction and traversal at this scale may introduce computational overheads, latency, and storage bottlenecks that our current evaluation does not capture. Addressing these scalability issues is an important direction for future work to ensure robustness when applied to substantially larger personal knowledge bases.

\bibliography{references}

\begin{thebibliography}{23}
\expandafter\ifx\csname natexlab\endcsname\relax\def\natexlab#1{#1}\fi

\bibitem[{Asai et~al.(2023)Asai, Wu, Wang, Sil, and Hajishirzi}]{asai2023selfraglearningretrievegenerate}
Akari Asai, Zeqiu Wu, Yizhong Wang, Avirup Sil, and Hannaneh Hajishirzi. 2023.
\newblock \href {http://arxiv.org/abs/2310.11511} {Self-rag: Learning to retrieve, generate, and critique through self-reflection}.

\bibitem[{Chen et~al.(2024)Chen, Wang, Li, Li, Yu, and Song}]{chen2024unifiedtemporalknowledgegraph}
Kai Chen, Ye~Wang, Yitong Li, Aiping Li, Han Yu, and Xin Song. 2024.
\newblock \href {http://arxiv.org/abs/2405.18106} {A unified temporal knowledge graph reasoning model towards interpolation and extrapolation}.

\bibitem[{Edge et~al.(2025)Edge, Trinh, Cheng, Bradley, Chao, Mody, Truitt, Metropolitansky, Ness, and Larson}]{edge2025localglobalgraphrag}
Darren Edge, Ha~Trinh, Newman Cheng, Joshua Bradley, Alex Chao, Apurva Mody, Steven Truitt, Dasha Metropolitansky, Robert~Osazuwa Ness, and Jonathan Larson. 2025.
\newblock \href {http://arxiv.org/abs/2404.16130} {From local to global: A graph rag approach to query-focused summarization}.

\bibitem[{Gao et~al.(2024)Gao, Xiong, Gao, Jia, Pan, Bi, Dai, Sun, Wang, and Wang}]{gao2024retrievalaugmentedgenerationlargelanguage}
Yunfan Gao, Yun Xiong, Xinyu Gao, Kangxiang Jia, Jinliu Pan, Yuxi Bi, Yi~Dai, Jiawei Sun, Meng Wang, and Haofen Wang. 2024.
\newblock \href {http://arxiv.org/abs/2312.10997} {Retrieval-augmented generation for large language models: A survey}.

\bibitem[{Guu et~al.(2020)Guu, Lee, Tung, Pasupat, and Chang}]{10.5555/3524938.3525306}
Kelvin Guu, Kenton Lee, Zora Tung, Panupong Pasupat, and Ming-Wei Chang. 2020.
\newblock Realm: retrieval-augmented language model pre-training.
\newblock In \emph{Proceedings of the 37th International Conference on Machine Learning}, ICML'20. JMLR.org.

\bibitem[{Hofer et~al.(2024)Hofer, Obraczka, Saeedi, Köpcke, and Rahm}]{Hofer_2024}
Marvin Hofer, Daniel Obraczka, Alieh Saeedi, Hanna Köpcke, and Erhard Rahm. 2024.
\newblock \href {https://doi.org/10.3390/info15080509} {Construction of knowledge graphs: Current state and challenges}.
\newblock \emph{Information}, 15(8):509.

\bibitem[{Hogan et~al.(2021)Hogan, Blomqvist, Cochez, D’amato, Melo, Gutierrez, Kirrane, Gayo, Navigli, Neumaier, Ngomo, Polleres, Rashid, Rula, Schmelzeisen, Sequeda, Staab, and Zimmermann}]{Hogan_2021}
Aidan Hogan, Eva Blomqvist, Michael Cochez, Claudia D’amato, Gerard~De Melo, Claudio Gutierrez, Sabrina Kirrane, José Emilio~Labra Gayo, Roberto Navigli, Sebastian Neumaier, Axel-Cyrille~Ngonga Ngomo, Axel Polleres, Sabbir~M. Rashid, Anisa Rula, Lukas Schmelzeisen, Juan Sequeda, Steffen Staab, and Antoine Zimmermann. 2021.
\newblock \href {https://doi.org/10.1145/3447772} {Knowledge graphs}.
\newblock \emph{ACM Computing Surveys}, 54(4):1–37.

\bibitem[{Hu et~al.(2024)Hu, Lei, Zhang, Pan, Ling, and Zhao}]{hu2024graggraphretrievalaugmentedgeneration}
Yuntong Hu, Zhihan Lei, Zheng Zhang, Bo~Pan, Chen Ling, and Liang Zhao. 2024.
\newblock \href {http://arxiv.org/abs/2405.16506} {Grag: Graph retrieval-augmented generation}.

\bibitem[{Huang et~al.(2025)Huang, Yu, Ma, Zhong, Feng, Wang, Chen, Peng, Feng, Qin, and Liu}]{10.1145/3703155}
Lei Huang, Weijiang Yu, Weitao Ma, Weihong Zhong, Zhangyin Feng, Haotian Wang, Qianglong Chen, Weihua Peng, Xiaocheng Feng, Bing Qin, and Ting Liu. 2025.
\newblock \href {https://doi.org/10.1145/3703155} {A survey on hallucination in large language models: Principles, taxonomy, challenges, and open questions}.
\newblock \emph{ACM Trans. Inf. Syst.}, 43(2).

\bibitem[{Irugalbandara et~al.(2024)Irugalbandara, Mahendra, Daynauth, Arachchige, Dantanarayana, Flautner, Tang, Kang, and Mars}]{10590016}
Chandra Irugalbandara, Ashish Mahendra, Roland Daynauth, Tharuka~Kasthuri Arachchige, Jayanaka Dantanarayana, Krisztian Flautner, Lingjia Tang, Yiping Kang, and Jason Mars. 2024.
\newblock \href {https://doi.org/10.1109/ISPASS61541.2024.00034} {Scaling down to scale up: A cost-benefit analysis of replacing openai's llm with open source slms in production}.
\newblock In \emph{2024 IEEE International Symposium on Performance Analysis of Systems and Software (ISPASS)}, pages 280--291.

\bibitem[{Jin et~al.(2024)Jin, Xie, Zhang, Roy, Zhang, Li, Li, Tang, Wang, Meng, and Han}]{jin2024graphchainofthoughtaugmentinglarge}
Bowen Jin, Chulin Xie, Jiawei Zhang, Kashob~Kumar Roy, Yu~Zhang, Zheng Li, Ruirui Li, Xianfeng Tang, Suhang Wang, Yu~Meng, and Jiawei Han. 2024.
\newblock \href {http://arxiv.org/abs/2404.07103} {Graph chain-of-thought: Augmenting large language models by reasoning on graphs}.

\bibitem[{Karpukhin et~al.(2020)Karpukhin, Oguz, Min, Lewis, Wu, Edunov, Chen, and Yih}]{karpukhin-etal-2020-dense}
Vladimir Karpukhin, Barlas Oguz, Sewon Min, Patrick Lewis, Ledell Wu, Sergey Edunov, Danqi Chen, and Wen-tau Yih. 2020.
\newblock \href {https://doi.org/10.18653/v1/2020.emnlp-main.550} {Dense passage retrieval for open-domain question answering}.
\newblock In \emph{Proceedings of the 2020 Conference on Empirical Methods in Natural Language Processing (EMNLP)}, pages 6769--6781, Online. Association for Computational Linguistics.

\bibitem[{Kim et~al.(2024)Kim, Fran{\c{c}}ois-Lavet, and Cochez}]{kim2024leveraging}
Taewoon Kim, Vincent Fran{\c{c}}ois-Lavet, and Michael Cochez. 2024.
\newblock Leveraging knowledge graph-based human-like memory systems to solve partially observable markov decision processes.
\newblock \emph{arXiv preprint arXiv:2408.05861}.

\bibitem[{Lewis et~al.(2021)Lewis, Perez, Piktus, Petroni, Karpukhin, Goyal, Küttler, Lewis, tau Yih, Rocktäschel, Riedel, and Kiela}]{lewis2021retrievalaugmentedgenerationknowledgeintensivenlp}
Patrick Lewis, Ethan Perez, Aleksandra Piktus, Fabio Petroni, Vladimir Karpukhin, Naman Goyal, Heinrich Küttler, Mike Lewis, Wen tau Yih, Tim Rocktäschel, Sebastian Riedel, and Douwe Kiela. 2021.
\newblock \href {http://arxiv.org/abs/2005.11401} {Retrieval-augmented generation for knowledge-intensive nlp tasks}.

\bibitem[{Niu et~al.(2024)Niu, Wu, Zhu, Xu, Shum, Zhong, Song, and Zhang}]{niu-etal-2024-ragtruth}
Cheng Niu, Yuanhao Wu, Juno Zhu, Siliang Xu, KaShun Shum, Randy Zhong, Juntong Song, and Tong Zhang. 2024.
\newblock \href {https://doi.org/10.18653/v1/2024.acl-long.585} {{RAGT}ruth: A hallucination corpus for developing trustworthy retrieval-augmented language models}.
\newblock In \emph{Proceedings of the 62nd Annual Meeting of the Association for Computational Linguistics (Volume 1: Long Papers)}, pages 10862--10878, Bangkok, Thailand. Association for Computational Linguistics.

\bibitem[{Peng et~al.(2024)Peng, Zhu, Liu, Bo, Shi, Hong, Zhang, and Tang}]{peng2024graphretrievalaugmentedgenerationsurvey}
Boci Peng, Yun Zhu, Yongchao Liu, Xiaohe Bo, Haizhou Shi, Chuntao Hong, Yan Zhang, and Siliang Tang. 2024.
\newblock \href {http://arxiv.org/abs/2408.08921} {Graph retrieval-augmented generation: A survey}.

\bibitem[{Qu et~al.(2024)Qu, Tu, and Bao}]{qu2024semanticchunkingworthcomputational}
Renyi Qu, Ruixuan Tu, and Forrest Bao. 2024.
\newblock \href {http://arxiv.org/abs/2410.13070} {Is semantic chunking worth the computational cost?}

\bibitem[{Su et~al.(2024)Su, Wang, Gao, Liu, Giunchiglia, Clevert, and Zitnik}]{su2024knowledgegraphbasedagent}
Xiaorui Su, Yibo Wang, Shanghua Gao, Xiaolong Liu, Valentina Giunchiglia, Djork-Arné Clevert, and Marinka Zitnik. 2024.
\newblock \href {http://arxiv.org/abs/2410.04660} {Knowledge graph based agent for complex, knowledge-intensive qa in medicine}.

\bibitem[{Sun et~al.(2025)Sun, Zang, Zheng, Song, Xu, Zhang, Yu, Song, and Li}]{sun2025redeepdetectinghallucinationretrievalaugmented}
Zhongxiang Sun, Xiaoxue Zang, Kai Zheng, Yang Song, Jun Xu, Xiao Zhang, Weijie Yu, Yang Song, and Han Li. 2025.
\newblock \href {http://arxiv.org/abs/2410.11414} {Redeep: Detecting hallucination in retrieval-augmented generation via mechanistic interpretability}.

\bibitem[{Wu et~al.(2024{\natexlab{a}})Wu, Xiong, Cui, Wu, Chen, Yuan, Huang, Liu, Kuo, Guan et~al.}]{wu2024retrieval}
Shangyu Wu, Ying Xiong, Yufei Cui, Haolun Wu, Can Chen, Ye~Yuan, Lianming Huang, Xue Liu, Tei-Wei Kuo, Nan Guan, et~al. 2024{\natexlab{a}}.
\newblock Retrieval-augmented generation for natural language processing: A survey.
\newblock \emph{arXiv preprint arXiv:2407.13193}.

\bibitem[{Wu et~al.(2024{\natexlab{b}})Wu, Huang, Hu, Hua, Qi, Chen, and Pan}]{Wu2024CoTKRCE}
Yike Wu, Yi~Huang, Nan Hu, Yuncheng Hua, Guilin Qi, Jiaoyan Chen, and Jeff~Z. Pan. 2024{\natexlab{b}}.
\newblock \href {https://api.semanticscholar.org/CorpusID:272986661} {Cotkr: Chain-of-thought enhanced knowledge rewriting for complex knowledge graph question answering}.
\newblock In \emph{Conference on Empirical Methods in Natural Language Processing}.

\bibitem[{Zhang et~al.(2024)Zhang, Zhang, Zhuang, Zhao, Wang, and Zheng}]{10504973}
Fuwei Zhang, Zhao Zhang, Fuzhen Zhuang, Yu~Zhao, Deqing Wang, and Hongwei Zheng. 2024.
\newblock \href {https://doi.org/10.1109/TKDE.2024.3390683} {Temporal knowledge graph reasoning with dynamic memory enhancement}.
\newblock \emph{IEEE Transactions on Knowledge and Data Engineering}, 36(11):7115--7128.

\bibitem[{Zhang et~al.(2022)Zhang, Beauchamp, and Wang}]{zhang2022generative}
Xinliang~Frederick Zhang, Nick Beauchamp, and Lu~Wang. 2022.
\newblock Generative entity-to-entity stance detection with knowledge graph augmentation.
\newblock \emph{arXiv preprint arXiv:2211.01467}.

\end{thebibliography}
\bibliographystyle{acl_natbib}

\end{document}